\documentclass[lettersize,journal]{IEEEtran}
\usepackage{amsmath,amsfonts}
\usepackage{algorithmic}
\usepackage{algorithm}
\usepackage{array}
\usepackage[caption=false,font=small,labelfont={rm,md,up},textfont={rm,md,up}]{subfig}
\usepackage{textcomp}
\usepackage{stfloats}
\usepackage{url}
\usepackage{verbatim}
\usepackage{graphicx}
\usepackage{cite}
\usepackage{amsmath,amssymb,amsfonts}
\usepackage{algorithmic}
\usepackage{graphicx}

\usepackage{comment}
\usepackage{multirow}
\usepackage{textcomp}
\usepackage{xcolor}
\usepackage[english]{babel}
\usepackage{csquotes}

\begin{document}
\title{Exploring DNN Robustness Against Adversarial Attacks Using Approximate Multipliers}

\author{Mohammad~Javad~Askarizadeh, Ebrahim~Farahmand, Jorge~Castro-Godínez, Ali~Mahani\\ Laura~Cabrera-Quirós, Carlos~Salazar-García
\thanks{Mohammad~Javad~Askarizadeh, Ebrahim~Farahmand, and Ali~Mahani are with the Shahid Bahonar University of Kerman, Kerman, Iran. Ali~Mahani is also with the York University, Toronto, Canada.}
\thanks{Jorge~Castro-Godínez, Laura~Cabrera-Quirós, and Carlos~Salazar-García are with the Instituto Tecnol\'ogico de Costa Rica, Cartago, Costa Rica.}

}

\markboth{}%
{Askarizadeh \MakeLowercase{\textit{et al.}}: Exploring DNN Robustness against Adversarial Attacks using Approximate Multipliers}


\maketitle

\begin{abstract}
Deep Neural Networks (DNNs) have advanced in many real-world applications, such as healthcare and autonomous driving. However, their high computational complexity and vulnerability to adversarial attacks are ongoing challenges. In this letter, approximate multipliers are used to explore DNN robustness improvement against adversarial attacks.
By uniformly replacing accurate multipliers for state-of-the-art approximate ones in DNN layer models, we explore the DNNs’ robustness against various adversarial attacks in a feasible time. Results show up to 7\% accuracy drop due to approximations when no attack is present while improving robust accuracy up to 10\% when attacks applied.
\end{abstract}

\begin{IEEEkeywords}
Approximate computing, deep learning, robustness, adversarial machine learning,
\end{IEEEkeywords}

\section{Introduction}
\IEEEPARstart{I}{n} recent years, DNNs have played a vital role in improving the service quality of applications, such as, autonomous driving~\cite{7995703}, healthcare~\cite{barata2019deep}, object detection~\cite{8417976}, and machine translation. DNNs face challenges in achieving acceptable accuracy in complex applications because of their immense computational and memory requirements~\cite{computation}, but also due to their vulnerability to adversarial attacks~\cite{goodfellow2014explaining}. In a nutshell, an adversarial attack is a noise introduced to the inputs. For instance, in computer vision applications a perturbed input may cause a misclassification by a DNN classifier~\cite{goodfellow2014explaining}. Enhancing the robustness of DNNs to adversarial attacks is crucial, particularly for safety-critical uses, such as autonomous driving~\cite{9207635}.

In the context of DNNs, Approximate Computing (AxC) aims to implement complex networks on limited resources, leveraging DNNs' inherent error resilience~\cite{venkataramani2020efficient, neggaz2018reliability, neggaz2019cnns} to enhance efficiency and performance at all layers. AxC techniques have been reported for enhancing DNN models’ robustness against adversarial attacks while maintaining effective trade-offs between energy savings and accuracy loss.

Some research in AxC has focused on proposing efficient approximate multipliers for DNNs with high computational requirements. DRUM is specially designed to have a scalable, configurable, and unbiased error distribution which is used for approximate applications~\cite{drum}. Farahmand \textit{et al}.~\cite{farahmand2023scaletrim} proposed scalable approximate multipliers using the linearization function and LUT-based compensation terms called scaleTRIM. The performance of their scaleTRIM configurations has been investigated with image classification using DNNs. The aforementioned state-of-the-art approximate multipliers have not been considered yet for assessing their impact on DNN model's robustness.

\vspace{8pt}
\noindent
\textbf{Related Work:} Reported works have proposed to consider the impact of AC techniques on the robustness of DNNs. EMPIR used an ensembling technique to create ensembles of mixed-precision DNN models~\cite{sen2020empir}. However, this approach led to a duplication of computing and memory units. The hardware implementation costs and execution time of complex DNNs, such as ResNet~\cite{resnet}, are excessively high. Sajadimanesh et al.~\cite{sajadimanesh2023eam} proposed an ensembling technique in which some DNN models are combined with exact and approximate compressors to enhance their robustness. This approach involves the replacement of exact compressors with approximate ones and their evaluations were performed on simple DNNs with a maximum of three convolution layers. Also, the approach uses an exact model for higher accuracy on benign inputs and an approximate model for higher robustness on perturbed inputs. This substitution does not yield a significant improvement in hardware implementation or energy efficiency because they used an exact model and also a model that uses only one type of approximate multiplier.

The design of an approximate multiplier unit aims to save computational resources such as area, delay, and power~\cite{Armeniakos2022}. However, the complexity of designing approximate arithmetic units and their impact on DNN accuracy makes it necessary for an approximate emulation framework. Popular DNN frameworks, such as Pytorch and TensorFlow, do not support approximate arithmetic, hence, using approximate arithmetic instead of accurate can slow down the emulation. AdaPT~\cite{danopoulos2022adapt} is a framework for fast cross-layer evaluation and retraining of DNN models based on the Pytorch library, and accelerates the process of DNN simulation by using multi-threading and vectorization. It supports any bit width and can be used for various DNN models. However, there is no option to explore against adversarial attacks.

In this work, we implement some state-of-the-art approximate multipliers in Adapt, to fast explore the accuracy and robustness behavior of different DNN models with approximate multipliers replaced by accurate ones in their layers, since the most power-consuming component of the Multiply-Accumulate (MAC) unit is the multiplier. To achieve this, we use the AdaPT framework. \textit{The novel contributions of this work are:}
\begin{itemize}
    \item Introduce newer state-of-the-art approximate multipliers into AdaPT framework.
    \item Introduce adversarial attacks in the AdaPT framework in order to robustness analysis of DNN models.
    \item Explore robustness and accuracy of complex CNN models, e.g. ResNet-50, under approximations and adversarial attacks in a feasible time.
\end{itemize}

\section{Robustness Improvement with Approximations}
\label{sectio_Robustness}
\begin{figure} [h]
    \centering
    \includegraphics[scale=0.5, width=1\linewidth]{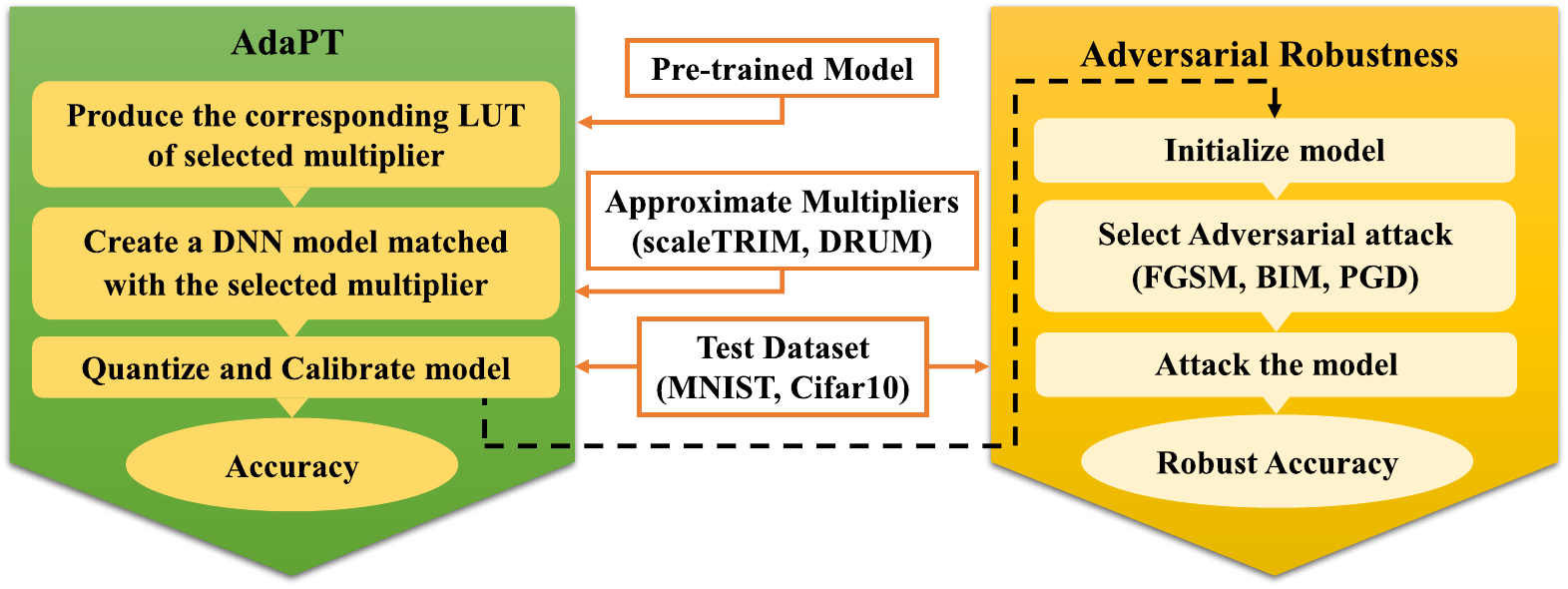}
    \caption{Overview of the proposed method.}
    \label{fig:1}
\end{figure}

In this work, we study the impact of using approximate multipliers for improving DNN models robustness against different adversarial attacks. We use different scaleTRIM approximate multiplier configurations for their beneficial features: scaleTRIM provides good performance and energy efficiency by truncating computational tasks, it can adjust truncation and error-compensation levels, allowing for a customized trade-off between accuracy and efficiency, and it has been proven effective in DNN-based image classification tasks, demonstrating its practical applicability~\cite{farahmand2023scaletrim}. Furthermore, we consider DRUM multipliers to compare them with scaleTRIM multipliers and generalize our proposed methodology with different configurations of both approximate multipliers. Fig.~\ref{fig:1} shows an overview of our proposed methodology.

\noindent
\textbf{AdaPT:}
The first step involves using the AdaPT framework. The user must provide the framework with two inputs: a pre-trained DNN model and a set of the aforementioned approximate multipliers. Next, AdaPT’s Look-up Table (LUT) generator produces the corresponding LUT for all different configurations of approximate multipliers which are implemented in Python. To avoid performing MAC operations in floating point, all parameters of the pre-trained model, such as weights, activations, and biases, quantize to 8-bit integer. After that, the quantized model then adjusts according to the selected approximate multiplier. This step includes replacing the selected accurate multiplications with approximate one across all model layers uniformly. This means all multiply operations of approximated layers perform using the selected approximate multiplier individually. In the first iteration, the algorithm selects the accurate multiplier, and in subsequent iterations, the algorithm assigns approximate multipliers one-by-one to the model layers. Finally, each model, which is created based on the selected multiplier, is evaluated in terms of accuracy and robustness on the validation dataset and a subset of the test dataset, respectively.

\noindent
\textbf{Adversarial Robustness:} In the second step, three white-box attacks are considered for the evaluation of adversarial robustness. These adversarial attacks are briefly described as follows:

\begin{itemize}
    \item \textit{Fast Gradient Method} (FGSM)~\cite{goodfellow2014explaining} is a basic but effective method that aims to perturb input samples by taking a single step in the direction of the sign of the loss function's gradient with respect to the input.
    \item \textit{Basic Iterative Method} (BIM) \cite{kurakin2016adversarial} is an iterative variant of FGSM that enhances the perturbation process by applying multiple small steps of perturbation.
    \item \textit{Projected Gradient Descend} (PGD) \cite{madry2017towards} is an iterative adversarial attack method designed to generate strong adversarial examples similar to BIM. It aims to find perturbed inputs that cause a machine-learning model to misclassify its predictions while staying within a certain perturbation budget.
\end{itemize}

In order to ensure the effectiveness of adversarial examples and added perturbations, it is crucial that they are visually imperceptible to humans. However, modeling human perception accurately can be challenging. To address this, researchers have put forward three metrics:

\begin{itemize}
    \item $L_0$: Counts the number of pixels with different values at corresponding positions in the two images (Original and perturbed image).
    \item $L_2$: Measures the Euclidean distance between the two images. 
    \item $L_\infty$: Measures the maximum difference between corresponding pixels in the two images.
\end{itemize}
These metrics approximate how humans perceive visual differences and help in assessing the impact of these perturbations accurately while maintaining their undetectable to the human eye \cite{carlini2017towards}. Eventually, the robust accuracy of DNN models is obtained under all mentioned attacks respectively. Multiple perturbation budgets, ranging from $0$ to $2$, are used for generating the adversarial examples using the test dataset. Higher perturbation budget values will increase the strength of the adversarial attack.

\section{Experimental Results}
To evaluate our proposed framework, in this letter, we explore the three mentioned adversarial attacks, FGSM, BIM, and PGD, on three DNN models which are implemented with six different configurations of approximate multipliers, from scaleTRIM and DRUM.

\subsection{DNN models and Datasets}
We consider three architectures for DNNs, LeNet-5~\cite{lenet}, ResNet-50~\cite{resnet}, and VGG-19~\cite{VGG} for two types of datasets, MNIST and CIFAR10. The LeNet-5 architecture was trained on the MNIST dataset, while CIFAR-10 dataset was used to train the ResNet-50 and VGG-19 architectures. All networks were trained and tested using Pytorch. We quantize all model parameters by using AdaPT framework features and use ReLU as an activation function. Table~\ref{tab:tab1} describes the structure of used networks in our exploration.

\begin{table}[h]
    \caption{Architecture of test DNNs}
    \centering
    \begin{tabular}{c c c c}
        \hline
        \textbf{DNN} & \textbf{Dataset} & \textbf{Network topology} &  \textbf{Params} \\
        \hline
        LeNet-5 & MNIST & 2 Conv, 3 FC & 44K\\
 
   
      
        ResNet-50 & CIFAR10 & 52 Conv, 1 FC& 23M\\
     
     
     
        VGG-19 & CIFAR10 & 16 Conv, 3 FC& 38M\\
        \hline
    \end{tabular}
    \label{tab:tab1}
    \vspace{-12pt}
\end{table}

\subsection{Adversarial attacks and metrics}
As mentioned, three adversarial attacks are considered in this work: FGSM, BIM, and PGD. Table~\ref{tab:attacks} shows the adversarial parameters for the attacks. We apply perturbation budgets ranging from $0$ to $2$ across all attacks with \(L_2\) and \(L_\infty\) distance metrics to assess the models' robustness.
\begin{table}[h]
    \caption{Adversarial Attack Parameters}
    \centering
    \begin{tabular}{c c c c}
        \hline
        \textbf{Parameters} & \textbf{FGSM} & \textbf{BIM} & \textbf{PGD} \\
        \hline
         \textbf{\(\alpha\)} & - & 0.2 & 0.01 \\
         \textit{\textbf{No. of iterations}} & - & 50 & 50 \\ 
        \hline
    \end{tabular}
    \label{tab:attacks}
    \vspace{-12pt}
\end{table}
\subsection{Impact of using approximate multipliers on models' accuracy}
Table~\ref{tab3} presents the accuracy of the aforementioned architectures when the accurate (ACC) and approximate multipliers (DRUM and scaleTRIM) uniformly assigned across all layers. Multipliers in both convolutional and fully connected layers of LeNet-5 are substituted with approximate ones. In other DNN models, only the convolution layers’ multipliers are replaced with approximate ones. The reason for this is that deeper DNN models have a smaller number of fully connected layers compared to convolution layers (based on Table~\ref{tab:tab1}). Additionally, the total MAC operations in fully connected layers are less than in convolution layers. The first row and the first column of Table~\ref{tab3} represent the accurate LeNet-5 model. The second row and the third column represent an approximate model of ResNet-50, in which DRUM-4 multiplier is used in its convolution layers, while fully connected layers use accurate multiplications. Using approximation in approximate multipliers results in a small decrease in accuracy compared to the accurate model. In all cases, the scaleTRIM(4,8) and DRUM-3 multipliers have the lowest and highest amount of accuracy loss compared to the accurate model. DRUM-3 has poor accuracy performance because it only considers 3 out of the 8 bits for the multiply operation.

\begin{table}[htbp]
\caption{Accuracy of different DNN models}
\begin{center}
\begin{tabular}{c c c c c c c c}
    \hline
    \multirow{2}{*}{\textbf{Models}} & \textbf{Accurate} & \multicolumn{2}{c}{\textbf{DRUM}} & \multicolumn{4}{c}{\textbf{scaleTRIM}} \\
    \cline{2-8}
 & \multirow{2}{*}{\textbf{ACC}}& \multirow{2}{*}{\textbf{DR3}}& \multirow{2}{*}{\textbf{DR4}} & \textbf{sT}& \textbf{sT}& \textbf{sT} & \textbf{sT} \\
 & & &  & \textbf{(3,0)}& \textbf{(4,0)}& \textbf{(4,4)} & \textbf{(4,8)} \\
\hline
LeNet-5 & \textbf{98.73} & 98.12 & 98.52& 98.60 & 98.68 & 98.62 & 98.66 \\



ResNet-50 & 93.54 & 86.16 & \textbf{91.94} & 92.48 & 93.38 & 93.4 & 93.42 \\



VGG-19 & 93.78 & 91.18 & 93.22& 93.16 & 93.63 & 93.66 & 93.79 \\
\hline
\end{tabular}
\label{tab3}
\end{center}
\vspace{-12pt}
\end{table}

\subsection{Impact of using of using approximate multiplier on model's robustness}
\begin{figure*} [ht]
    \centering
        \subfloat[{\footnotesize Robust accuracy vs Perturbation budegt in LeNet-5}\label{fig:l5}]{%
          \includegraphics[width=0.3\textwidth]{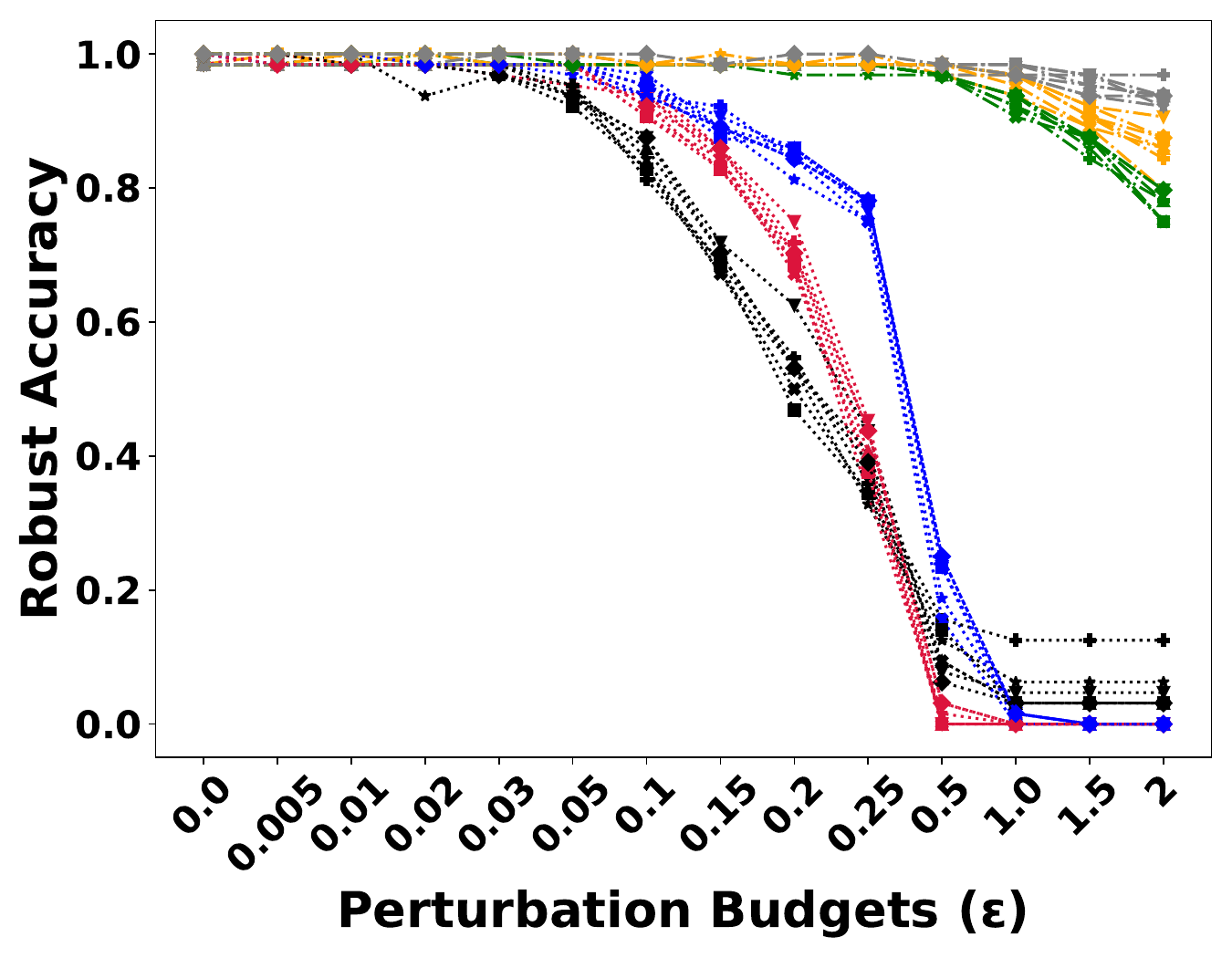}
        }
        \qquad
        \subfloat[{\footnotesize Robust accuracy vs Perturbation budegt in ResNet-50}\label{fig:r50}]{%
          \includegraphics[width=0.3\textwidth]{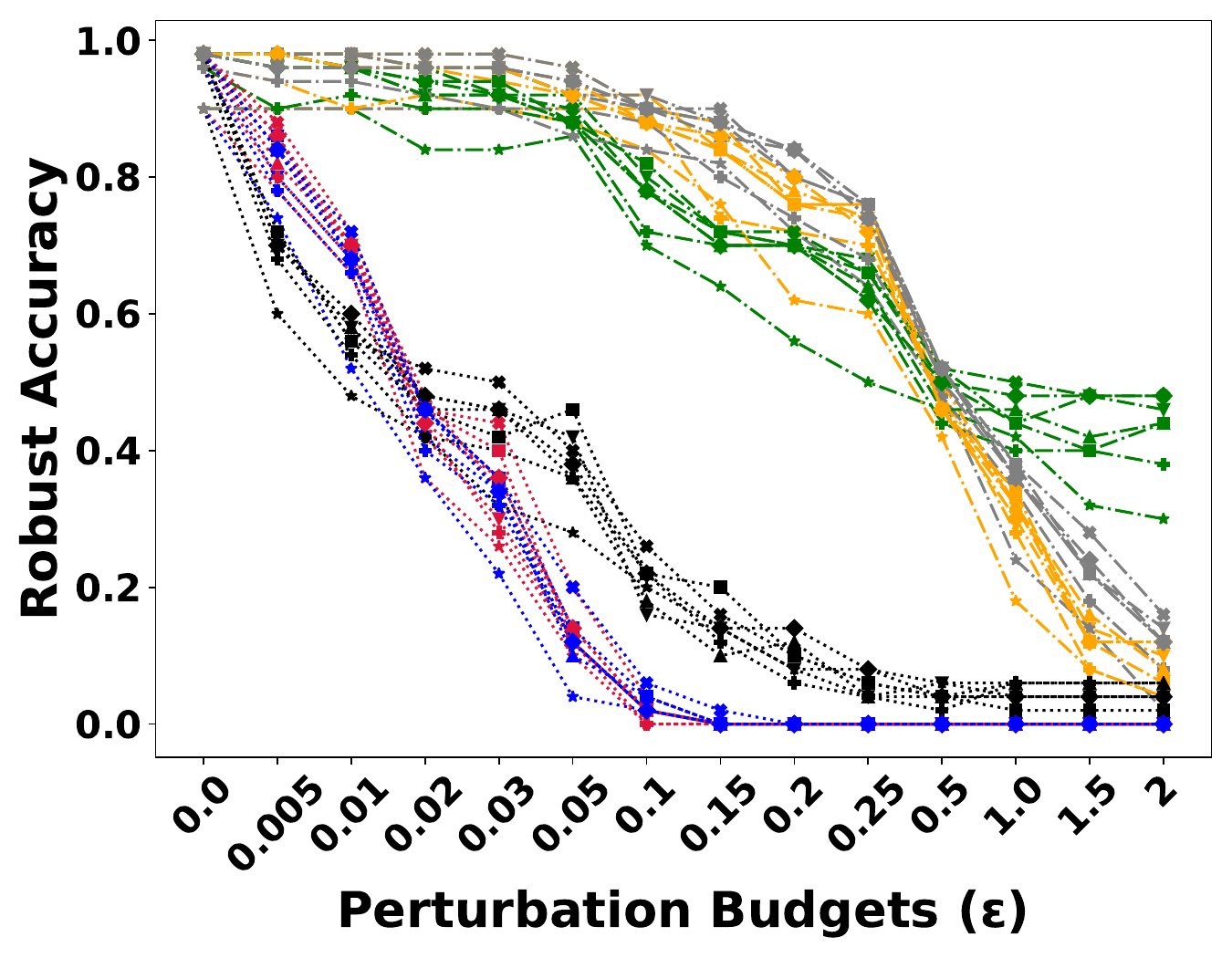}
        }
        \qquad
        \subfloat[{\footnotesize Robust accuracy vs Perturbation budegt in VGG-19}\label{fig:v19}]{%
          \includegraphics[width=0.3\textwidth]{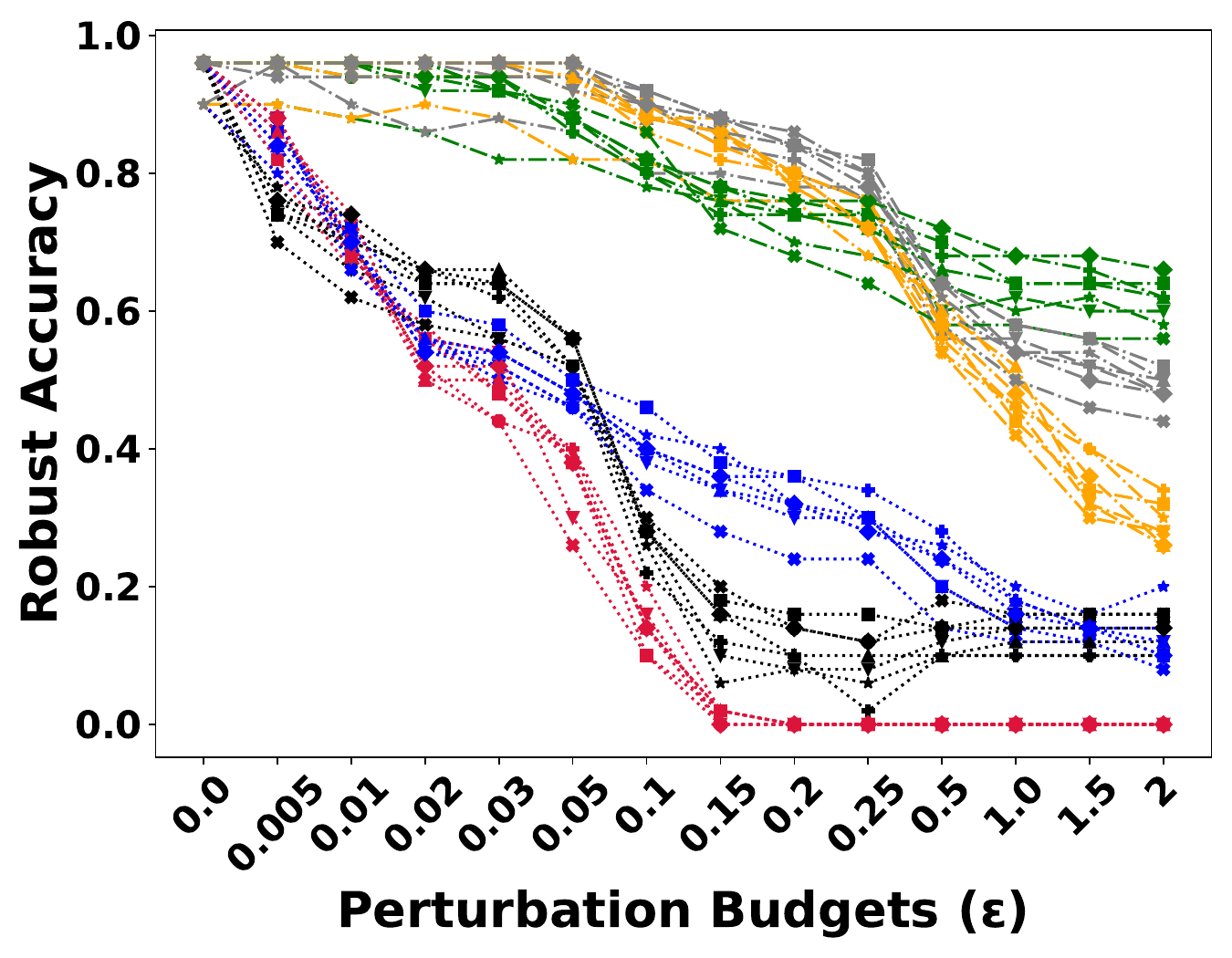}
        }
    \caption{Exploration of DNN models' robust accuracy against adversarial attack, (a) LeNet-5, (b)  ResNet-50, and (c) VGG-19}
    \label{fig:2}
    \vspace{-12pt}
\end{figure*}

For assessing robustness of accurate and approximate models, their classification accuracy is evaluated on adversarial images created using different attacks. The classification accuracy under \(L_2\) and \(L_\infty\) of FGSM, BIM, and PGD attacks are reported for different perturbation budgets (\(\epsilon\)) ranging from $0$ to $2$ in Figure~\ref{fig:2}. These figures depict the results on a subset of test dataset images from the MNIST for LeNet-5 and the CIFAR-10 for the ResNet-50 and VGG-19 models.

Figure~\ref{fig:l5} shows the robust accuracy of LeNet-5 models in which multipliers are uniformly assigned to their layers. The robust accuracy performance of the approximate models can be compared to the accurate model in three ranges. For the LeNet-5 models, these ranges can be specified against the $L_{\infty}$ attack as (1): $0 < \epsilon < 0.02$, (2): $0.02 \leq \epsilon < 0.15$, and (3): $0.15 \leq \epsilon < 2$. In the first region of first perturbation budgets, the robust accuracy of the approximate models is better than the accurate model and has been improved by 1.5625\% for, e.g., the sT(4,4) model at $\epsilon = 0.005$ under PGD and, the sT(4,8) model at $\epsilon = 0.005$ and 0.01. Also, in the second range, the robust accuracy has been improved by 1.5625\% rather than accurate model for, e.g., the DR4 model at $\epsilon = 0.1$ under PGD and, the sT(3,0) under BIM and the sT(3,0), sT(4,0) and sT(4,4) at $\epsilon = 0.3$ under FGSM attack. For the third range, none of the approximate models exhibits a better performance than the accurate model. The DR3 model, e.g., is the worst one at $\epsilon = 0.25$ with a 10.9375\% accuracy drop in PGD and FGSM attacks, and in BIM attack the sT(3,0) has the highest accuracy drop at $\epsilon = 0.5$ with a 7.8125\%. In the $L_2$ attacks, all approximate models either exhibit the same performance of robust accuracy as the accurate model.

Figure~\ref{fig:r50} illustrates the results for the ResNet-50, in which approximate multipliers are uniformly assigned to their convolution layers. DR3 models have the worst robust accuracy compared to the accurate model with a maximum 30\% accuracy drop at $\epsilon = 0.005$ and $\epsilon = 0.1$ against both $L_\infty$ and $L_2$ attacks, respectively. The approximate models with scaleTRIM have shown robust accuracy values near the accurate model. For example, across all $L_\infty$ attacks, the sT(3,0) has the most robust accuracy improvement at $\epsilon = 0.02$ by 8\% under PGD attack, and the sT(4,4) has the lowest robust accuracy decrement by 8\% under compared to the accurate model. Across all $L_2$ attacks, the highest increment and lowest decrement of approximate models' robust accuracy compared with the accurate model is belong to the sT(3,0) at $\epsilon = 1$ by 10\% under BIM and sT(4,4) at $\epsilon = 1$ under FGSM attack, respectively.

\begin{figure*} [!ht]
    \centering
    \subfloat{%
      \includegraphics[width=0.45\textwidth]{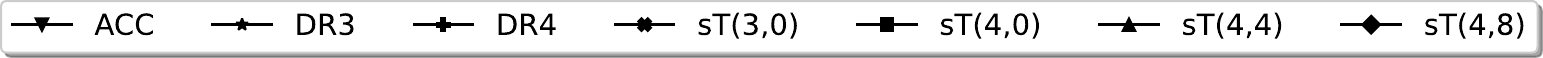}
    }
    \qquad
    \subfloat{%
      \includegraphics[width=0.45\textwidth]{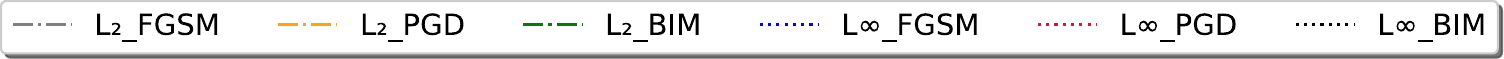}
      
    }     
\end{figure*}

Figure~\ref{fig:v19} depicts the robust accuracy of the VGG-19 against different adversarial attacks. For $L_\infty$ attacks, the DR4 model experiences a maximum accuracy drop of 10\% at $\epsilon = 0.03$ under PGD, while in $L_2$ attacks, the maximum accuracy drop is 10\% at different $\epsilon$ values of ranging from 0.02 to 0.15. The accuracy of models with scaleTRIM is compromising, comparable to that of the accurate model. For instance, when subjected to $L_\infty$ attacks, the sT(4,4) model showed an improvement in robust accuracy of 10\% at $\epsilon = 0.03$ under FGSM attack. In contrast, the sT(3,0) model showed the smallest drop in robust accuracy, which was also 10\% at $\epsilon = 0.05$ under PGD, when compared to the accurate model. In $L_2$ attacks, the sT(4,8) model has the highest increase by 12\% in robust accuracy compared to the accurate model, at $\epsilon = 0.5$ under the FGSM attack and the sT(3,0) model has the highest decrease in robust accuracy by 10\% at $\epsilon = 0.25$ under the FGSM attack.


\subsection{Time elapsed}
Table \ref{time} presents the overall time to evaluate for each and all models in terms of robustness and accuracy evaluations. All simulations are performed on Intel i7-1185G7 4.8 GHz (4 cores, 8 threads) with 16GB DDR4 RAM. The required time is feasible because of the utilization of the AdaPT emulator for performing the calculations. For instance, evaluations of ResNet-50 take 4 days, and all evaluations were completed within 14 days.
\begin{table}[htbp]
\caption{Time requirement to perform the evaluations}
\begin{center}
\begin{tabular}{cccc}
\hline
\textbf{Time elapsed}&\textbf{LeNet-5}&\textbf{ResNet-50}& \textbf{VGG-19} \\
\hline
1 eval. (s) & 14.74	&87.6&55.82\\
total (h) & 17.2&102.25&65.12 \\
\hline
\end{tabular}
\label{time}
\end{center}
\vspace{-12pt}
\end{table}

\section{Conclusion}

In this letter, we replaced accurate multipliers with approximate multipliers in DNN models to improve their robustness against adversarial attacks. Results shown that our approach could improve the robust accuracy by 10\% despite a maximum 7\% decrease in accuracy.
In the future, our plan is to develop AdaPT to automate finding the appropriate approximate multiplier and determining the level of approximation needed for each model layer. This will help us create optimal models that are both robust and accurate.

\bibliographystyle{IEEEtran}
\bibliography{new}
\end{document}